\documentclass[letterpaper]{article} 
\usepackage[camera-ready]{aaai24}  
\usepackage{times}  
\usepackage{helvet}  
\usepackage{courier}  
\usepackage[hyphens]{url}  
\usepackage{graphicx} 
\usepackage[square, numbers]{natbib}  
\usepackage{caption} 
\usepackage{algorithm}
\usepackage{algorithmic}
\usepackage{amsmath}
\usepackage{booktabs}
\usepackage{makecell}
\usepackage{booktabs}

\usepackage{bm}
\usepackage{tikz}
\usepackage{scalerel}
\usepackage{xspace}
\usepackage{pifont}
\usepackage{multirow}
\usepackage{tikz}
\usepackage{comment}
\usepackage{amsmath,amssymb} 
\usepackage{color}
\usepackage[inline]{enumitem}
\usepackage{float}
\usepackage{booktabs}       
\usepackage{indentfirst}
\usepackage{multirow}
\usepackage{makecell}
\usepackage{pifont}
\usepackage{everyshi}
\newcommand\doubleplus{+\kern-1.3ex+\kern0.8ex}

\newcommand{\cmark}{\ding{51}}%
\usepackage{colortbl}
\usepackage{marvosym}

\usepackage{newfloat}
\usepackage{listings}
\DeclareCaptionStyle{ruled}{labelfont=normalfont,labelsep=colon,strut=off} 
\floatstyle{ruled}
\newfloat{listing}{tb}{lst}{}
\floatname{listing}{Listing}
\title{SwitchTab: Switched Autoencoders Are Effective Tabular Learners}
\author{
    Jing Wu\equalcontrib,
    Suiyao Chen\equalcontrib,
    Qi Zhao,\\
    Renat Sergazinov,
    Chen Li,
    Shengjie Liu,
    Chongchao Zhao,
    Tianpei Xie,
    Hanqing Guo,
    Cheng Ji,
    Daniel Cociorva,
    Hakan Brunzell
}
\affiliations{
   Amazon Buyer Risk Prevention\\

    4575 La Jolla Village Dr\\
    San Diego, California 92122 USA\\
    \{jingwua, suiyaoc, qqzhao, renserg, chenlii, zycjlsj, zchongch, lukexie, hanqiguo, cjiamzn, cociorva, brunzell\}@amazon.com
}

\begin{document}
\maketitle
\begin{abstract}
Self-supervised representation learning methods have achieved significant success in computer vision and natural language processing, where data samples exhibit explicit spatial or semantic dependencies. However, applying these methods to tabular data is challenging due to the less pronounced dependencies among data samples. In this paper, we address this limitation by introducing SwitchTab, a novel self-supervised method specifically designed to capture latent dependencies in tabular data. SwitchTab leverages an asymmetric encoder-decoder framework to decouple mutual and salient features among data pairs, resulting in more representative embeddings. These embeddings, in turn, contribute to better decision boundaries and lead to improved results in downstream tasks. To validate the effectiveness of SwitchTab, we conduct extensive experiments across various domains involving tabular data. The results showcase superior performance in end-to-end prediction tasks with fine-tuning. Moreover, we demonstrate that pre-trained salient embeddings can be utilized as plug-and-play features to enhance the performance of various traditional classification methods (e.g., Logistic Regression, XGBoost, etc.). Lastly, we highlight the capability of SwitchTab to create explainable representations through visualization of decoupled mutual and salient features in the latent space. 
\end{abstract}

\begin{figure}[t]
\begin{center}
    \includegraphics[width=0.9\linewidth]{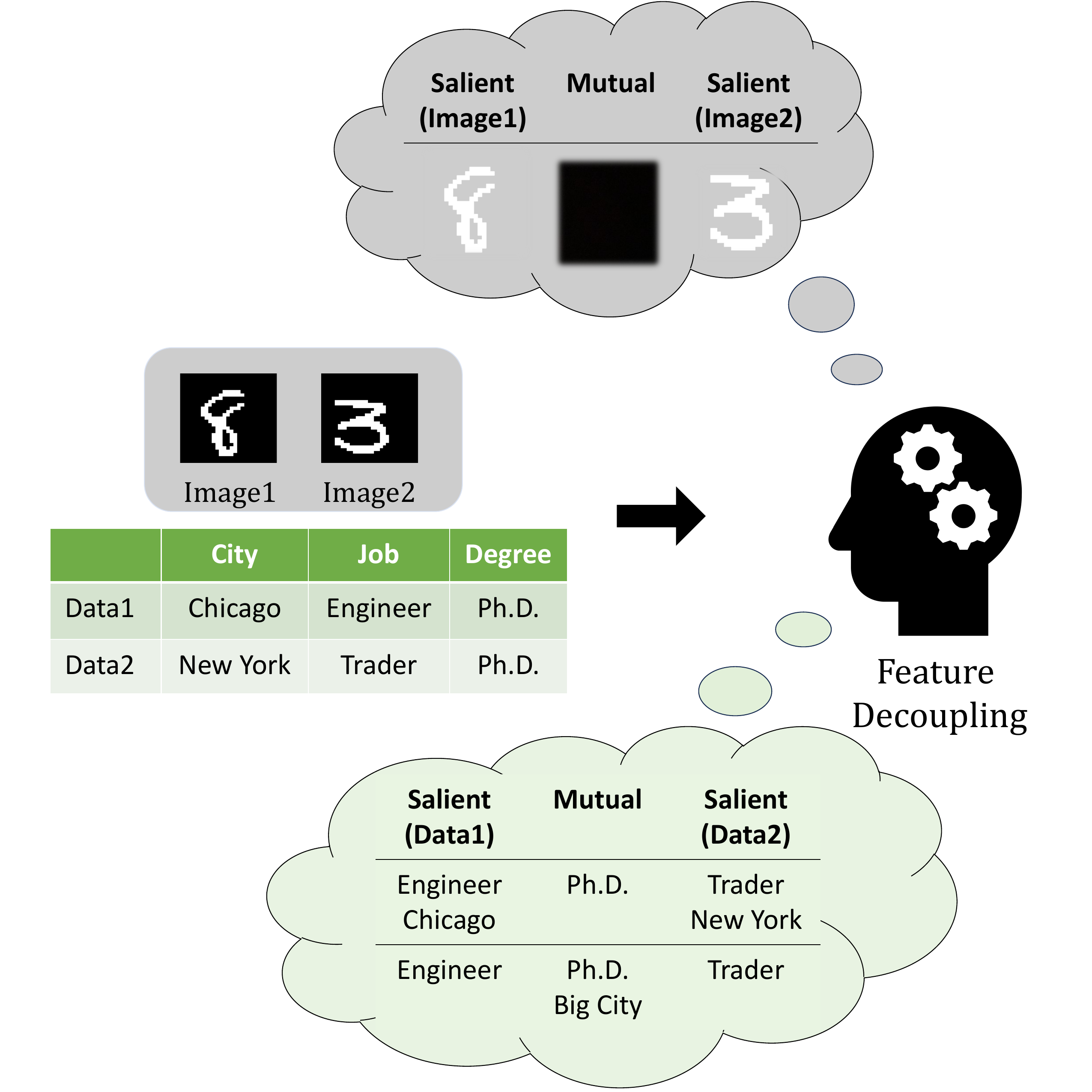}
\end{center}
  \vspace*{-3mm} 
  \caption{Given a pair of images, a person can easily distinguish the salient digits and mutual background due to the well-structured spatial relationships. However, it becomes challenging to distinguish a pair of tabular samples. For instance, feature City may be salient between data points ``Chicago" and ``New York" for word counts, however, still sharing some latent mutual information (e.g., big cities), making it challenging for decoupling. Note that this decoupling process is for illustration only. In the implementation, all the decoupled samples are computed in the feature space.}
  \vspace*{-3mm} 
\label{fig: demo}
\end{figure}

\section{Introduction}

While representation learning \cite{bengio2013representation} has made remarkable advancements in computer vision (CV) and natural language processing (NLP) domains, tabular data, which is ubiquitous in real-world applications and critical industries such as healthcare \cite{qayyum2020secure,chen2017personalized, dong2021semi, osorio2022executive, dong2021using, tang2023ours}, manufacturing \cite{borisov2022deep, chen2017multi,bingjie2023optimal,chen2020optimal, dong2023graph}, agriculture \cite{liakos2018machine, wu2022optimizing, tao2022optimizing}, transportation \cite{li2023context,zhang2023empirical,zhang2020hybrid, dong2022utility} and various engineering fields \cite{zhu2018big,chen2020some, wang2023balanced,wang2019inverse,gao2023autonomous, ye2023demultiplexing}, has not fully benefited from its transformative power and remains relatively unexplored. Traditionally, researchers in these domains leverage domain expertise for feature selection \citep{covert2019deep}, model refinement \citep{wang2018surrogate,wang2023scalable,wang2023ntk,wang2023lemon} and uncertainty quantification \citep{chen2018data, wang2023inverse, wang2019gaussian}. The unique challenges posed by tabular datasets stem from their inherent heterogeneity, which lacks explicit spatial relationships in images (e.g., similar background and distinct characters)  or semantic dependencies in languages. Tabular data typically comprises redundant features that are both numerical and categorical, exhibiting various discrete and continuous distributions \cite{grinsztajn2022tree}. These features can be either dependent or entirely independent from each other, making it difficult for representation learning models to capture crucial latent features for effective decision-making or accurate predictions across diverse samples. 

When comparing data samples, mutual features consist of information that highlights common characteristics, while salient features emphasize the distinctive attributes to differentiate one sample from the others. For image data, the intensity of the background pixels forms the mutual features shared across images while the relative positions of bright and dark pixels form the salient features, which are likely to vary significantly across images with different shapes or objects. As illustrated in Figure~\ref{fig: demo}, in MNIST \cite{xiao2017fashion}, decoupling digits from the background is relatively straightforward, using digits as the salient features for classification. However, the differentiation for tabular data tends to be less distinct. For example, feature like City can be considered salient when the data points ``Chicago" and ``New York" have different word counts. Nonetheless, when considering the size of city semantically, feature City could share mutual information. Therefore, it becomes more complicated to set the decision boundary for classification. 

To tackle these challenges, our central insight revolves around empowering representation models to explicitly distinguish mutual and salient information within the feature space, which we define as the decoupling process. Instead of solely relying on the original data space, We firmly believe that manipulating the feature space could lead to less noise and obtain more representativeness, adapting the success of representation learning from other domains to tabular data.

In this paper, we introduce SwitchTab, an elegant and effective generative pre-training framework for tabular data representation learning. The core of SwitchTab is an asymmetric encoder-decoder structure, augmented with custom projectors that facilitate information decoupling. The process begins with encoding each data sample into a general embedding, which is further projected into salient and mutual embeddings. What sets SwitchTab apart is the deliberate swapping of salient and mutual embeddings among different data samples during decoding. This innovative approach not only allows the model to acquire more structured embeddings from encoder but also explicitly extracts and represents the salient and mutual information. Another advantage of SwitchTab is its versatility, to be trained effectively in both self-supervised manners. This adaptability ensures that SwitchTab performs well in diverse training scenarios, regardless of the availability of labeled data.

Our contributions can be summarized as follows:

\begin{itemize}[noitemsep,topsep=0pt]

    \item[$\bullet$]We propose SwitchTab, a novel self-supervised learning framework to decouple salient and mutual embeddings across data samples. To the best of our knowledge, this is the first attempt to explore and explicitly extract separable and organized embeddings for tabular data. 

    \item[$\bullet$]By fine-tuning the pre-trained encoder from SwitchTab, we demonstrate that our method achieves competitive results across extensive datasets and benchmarks.

    \item[$\bullet$]The extracted salient embeddings can be used as plug-and-play features to enhance the performance of various traditional prediction models, e.g., XGBoost.

    \item[$\bullet$]We visualize the structured embeddings learned from SwitchTab and highlight the distinction between mutual and salient information, enhancing the explainability of the proposed framework.

\end{itemize}

\section{Related Work}
\subsection{Models for Tabular Data Learning and Prediction}
\subsubsection{Traditional Models.} For tabular data classification and regression tasks, various machine learning methods have been developed. For linear relationships modeling, Logistic Regression (LR) \cite{wright1995logistic,zhang2020freeway} and Generalized Linear Models (GLM) \cite{hastie2017generalized,chen2019claims} are top choices. Tree-based models include Decision Trees (DT) \cite{breiman2017classification} and various ensemble methods based on DT such as XGBoost \cite{chen2016xgboost}, Random Forest \cite{breiman2001random}, CatBoost \cite{prokhorenkova2018catboost} and LightGBM \cite{ke2017lightgbm}, which are widely adopted in industry for modeling complex non-linear relationships, improving interpretability and handling various feature types like null values or categorical features. 

\subsubsection{Deep Learning Models.} Recent research trends aim to adopt deep learning models to tabular data domain. Various neural architectures have been introduced to improve performance on tabular data. There are several major categories \cite{borisov2022deep, gorishniy2021revisiting}, including 1) supervised methods with neural networks (e.g., ResNet \cite{he2016deep}, SNN \cite{klambauer2017self}, AutoInt \cite{song2019autoint}, DCN V2 \cite{wang2021dcn}); 2) hybrid methods to integrate decision trees with neural networks for end-to-end training (e.g., NODE \cite{popov2019neural}, GrowNet \cite{badirli2020gradient}, TabNN \cite{ke2018tabnn}, DeepGBM \cite{ke2019deepgbm}); 3) transformer-based methods to learn from attentions across features and data samples (e.g., TabNet \cite{arik2021tabnet}, TabTransformer \cite{huang2020tabtransformer}, FT-Transformer \cite{gorishniy2021revisiting}); and 4) representation learning methods, which have emerging focuses and align with the scope of our proposed work, to realize effective information extraction through self- and semi-supervised learning (e.g., VIME \cite{yoon2020vime}, SCARF \cite{bahri2021scarf}, SAINT \cite{somepalli2021saint}) and Recontab\cite{chen2023recontab}. In addition, indirect tabular learning would transfer tabular data into graph \cite{peng2023maxk, peng2023lingcn, xie2023accel} and define graph-related tasks for representation learning, such as TabGNN \cite{guo2021tabgnn}.

\begin{figure*}[t!]
\begin{center}
    \includegraphics[width=1\linewidth]{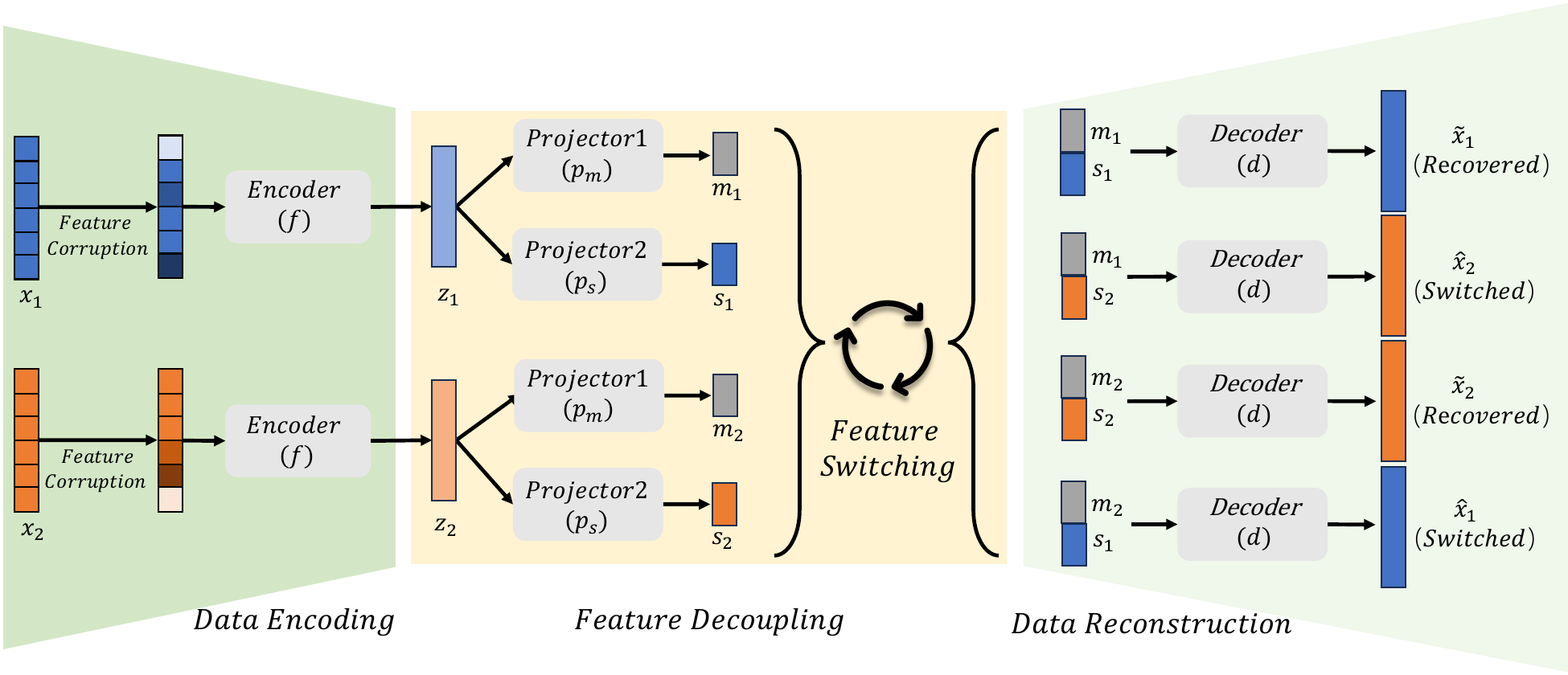}
\end{center}
  \caption{Block diagram of the proposed self-supervised learning framework. (1) Two different samples $x_1$ and $x_2$ are randomly corrupted and encoded into feature vectors $z_1$ and $z_2$ through encoder $f$. (2) feature vectors $z_1$ and $z_2$ are decoupled into mutual and salient features by two different projectors $p_m$ and $p_s$, respectively. (3) Mutual and salient features are combined and reconstructed by a decoder $d$ where the salient feature dominates the sample type and the mutual feature provides common information that is switchable among two samples.}
  \vspace*{-3mm} 
\label{fig: ssl}
\end{figure*}

\subsection{Self-supervised Representation Learning} Deep representation learning methods have been introduced in the computer vision and remote sensing domains, utilizing self-supervised learning methods \cite{kolesnikov2019revisiting, ericsson2022self, li2015semi,wu2023hallucination, manas2021seasonal, wu2023genco}. These methods can be divided into two branches. The first branch mainly focuses on a contrastive learning framework with various data augmentation schemes such as artificial augmentations, Fourier transform, or temporal variances\cite{nachum2021provable, wu2023extended, ye2023multiplexed,ye2023oam,ye2023free}. More specifically, models rely on momentum-update strategies \cite{he2020momentum, wu2023extended,chen2020improved,wu2023genco, che2023enhancing}, large batch sizes \cite{chen2020simple}, stop-gradient operations \cite{chen2021exploring}, or training an online network to predict the output of the target network \cite{grill2020bootstrap}. These ideas have also been applied to the tabular data domain. One representative work in this area is SCARF \cite{bahri2021scarf}, which adopts the idea of SimCLR \cite{chen2020simple} to pre-train the encoder using feature corruption as the data augmentation method. Another work is SAINT \cite{somepalli2021saint}, which also stems from a contrastive learning framework and computes column-wise and row-wise attentions.
The second branch is based on generative models such as autoencoders \cite{kingma2013auto}. Specifically, Masked Autoencoder (MAE) \cite{he2022masked} has an asymmetric encoder-decoder architecture for learning embeddings from images. This framework is also capable of capturing spatiotemporal information \cite{feichtenhofer2022masked} and can be extended to 3D space \cite{jiang2022masked} and multiple scales \cite{reed2022scale}. The similar masking strategy is widely used in NLP \cite{devlin2018bert} as well as tabular data \cite{arik2021tabnet, huang2020tabtransformer, yin2020tabert}. A work similar to MAE in the domain of tabular data is VIME \cite{yoon2020vime}. VIME corrupts and encodes each sample in feature space using two estimators. After each estimator, the features are assigned with decoders to reconstruct a binary mask and the original uncorrupted samples, respectively. The key difference between VIME and our work is that we leverage the asymmetric encoder-decoder architecture in pre-training \cite{chen2023recontab} and introduce a switching mechanism, which strongly encourages the encoder to generate more structured and representative embeddings.

\begin{algorithm}[t]
\caption{Self-supervised Learning with SwitchTab}
\label{algo:DA}
\begin{algorithmic}[1]
{\small
\REQUIRE unlabeled data  $\mathcal{X} \subseteq \mathbb{R}^M$, batch size $B$, encoder $f$, projector for mutual information $p_{m}$, projector for salient information $p_{s}$, decoder $d$, mean squared error {MSE}, feature concatenation $\oplus$.

\FOR {two sampled mini-batch $ \left\{x_{i}^{1}\right\}_{i=1}^{B} \subseteq \mathcal{X}$ and $ \left\{x_{i}^{2}\right\}_{i=1}^{B} \subseteq \mathcal{X}$} 
    \STATE for each sample $x_{i}^{1}$ and $x_{i}^{2}$, apply feature corruption, define the corrupted feature as: 
    $\breve{x}_{i}^{1}$ and $\breve{x}_{i}^{2}$, for $i \in [B]$
    \STATE data encoding:  
    
    $z_{i}^{1}=f(\breve{x}_{i}^{1})$, $z_{i}^{2}=f(\breve{x}_{i}^{2})$, for $i \in [B]$
    \STATE feature decoupling:
    
    (1) the salient and mutual information of the first batch be defined as follows: $s^{1}_{i}= p_s(z_{i}^{1})$ and $m^{1}_{i}= p_m(z_{i}^{1})$. 

    (2) the salient and mutual information of the second batch be defined as follows: $s^{2}_{i}= p_s(z_{i}^{2})$ and $m^{2}_{i}= p_m(z_{i}^{2})$.
    
    \STATE data reconstruction:
    
    (1) let recovered pairs be defined as:

    $\tilde{x}^{1}_{i} = d(m_{i}^{1}\oplus s_{i}^{1})$, $\tilde{x}^{2}_{i} = d(m_{i}^{2}\oplus s_{i}^{2})$ 

    (2) let switched pairs be defined as:
    
    $\hat{x}^{1}_{i} = d(m_{i}^{2}\oplus s_{i}^{1})$, $\hat{x}^{2}_{i} = d(m_{i}^{1}\oplus s_{i}^{2})$ 

    \STATE define reconstruction loss $\mathcal{L}_{recon}=$
    
    MSE$({x}^{1}_{i}, \tilde{x}^{1}_{i}) + $MSE$({x}^{2}_{i}, \tilde{x}^{2}_{i})+$MSE$({x}^{1}_{i}, \hat{x}^{1}_{i})+$MSE$({x}^{2}_{i}, \hat{x}^{2}_{i})$

    \STATE update encoder $f$, projectors $p_{m}$ and $p_{s}$, and decoder $d$ to minimize $\mathcal{L}_{recon}$ using RMSProp.

\ENDFOR
}
\end{algorithmic}
\end{algorithm}
\vspace*{-2mm}

\subsection{Feature Decoupling}
In the areas of feature extraction \cite{salau2019feature, qiao2023relation, zhou2023improving} and latent representation learning \cite{bengio2013representation}, autoencoder-based models \cite{kingma2013auto, abukmeil2021survey} have been widely used for , with strong capabilities to learning useful representations for real-world tasks with little or no supervision. Previous work has been focusing on learning a decoupled representation \cite{higgins2016beta, kim2018disentangling, bousmalis2016domain, zhang2020fdn} where each dimension can capture the change of one semantically meaningful factor of variation while being relatively invariant to changes in other factors. Recent work also explored capturing the dependencies and relationships across different factors of variation to enhance the latent representations \cite{sonderby2016ladder, tschannen2018recent}. Taking one step further, the work of contrastive variational autoencoder (cVAE) by \cite{Abid2019ContrastiveVA}, which adapted the contrastive analysis principles, has explicitly categorized latent features by salient and mutual information and enhanced the salient features. The swapping autoencoder by \cite{park2020swapping} explicitly decouple the image into structure and texture embeddings, which are swapped for image generation. Some recent work for tabular data representation learning has also shown the benefits of quantifying the between-sample relationships. Relational Autoencoder (RAE) \cite{meng2017relational} considered both the data features and relationships to generate more robust features with lower reconstruction loss and better performance in downstream tasks. \cite{kossen2021self,somepalli2021saint} shared a similar idea to consider self-attention between data samples. We extend the idea of cVAE and swapping autoencoder to the tabular data domain with the argument that the two data samples share mutual and salient information through latent between-sample relationships. Salient information is crucial for downstream tasks involving decision boundaries, while mutual information remains necessary for data reconstruction. To the best of our knowledge, we are the first to model tabular data with explicit and expressive feature decoupling architecture to enhance the representation learning performance. Meanwhile, feature decoupling could enhance the explainability of the model. Existing work has explored different perspectives such as SHAP value \cite{jethani2022fastshap}, concepts \cite{zarlenga2023tabcbm}, and counterfactual explanations \cite{chen2022relax,chen2022explain,chen2022grease}, etc. However, explicit learning of salient and mutual information from model structure is yet to be explored.

\section{Method}
In this section, we present SwitchTab, our comprehensive approach for tabular data representation learning and feature decoupling. First, we outline the process of feature corruption. Then, in the second sub-section, we delve into the intricacies of self-supervised learning, including data encoding, feature decoupling, and data reconstruction. The third sub-section elucidates our pre-training learning method with labels. Finally, we illustrate how to utilize the pre-trained encoders and embeddings to improve downstream tasks.

\subsection{Feature Corruption}\
\label{sec: corruption}
Generative-based representation learning relies on data augmentations to learn robust embeddings for downstream tasks. Among different methods, feature corruption \cite{yoon2020vime, bahri2021scarf} is one of the most promising approaches. In this paper, we also take advantage of this method to improve the model performance. For one tabular data $x_{i}$ from original dataset $\mathcal{X} \subseteq \mathbb{R}^M$, we define its $j$-th feature as $x_{i_j}$, i.e., $x_{i} = (x_{i_1}, x_{i_2},..., x_{i_M})$, where $M$ is the dimension of features and $i$ is the index of samples. For each sample, we randomly select $t$ features among $M$ features and replace them with corrupted feature $c$. Concretely, $c  \sim \widehat{\mathcal{X}}_{i_j}$, where $\widehat{\mathcal{X}}_{i_j}$ is the uniform distribution over $\mathcal{X}_{i_j} = \left\{ x_{i_j}: x_{i} \in \mathcal{X}\right\}$.

\begin{figure*}[t!]
\begin{center}
    \includegraphics[width=0.85\linewidth]{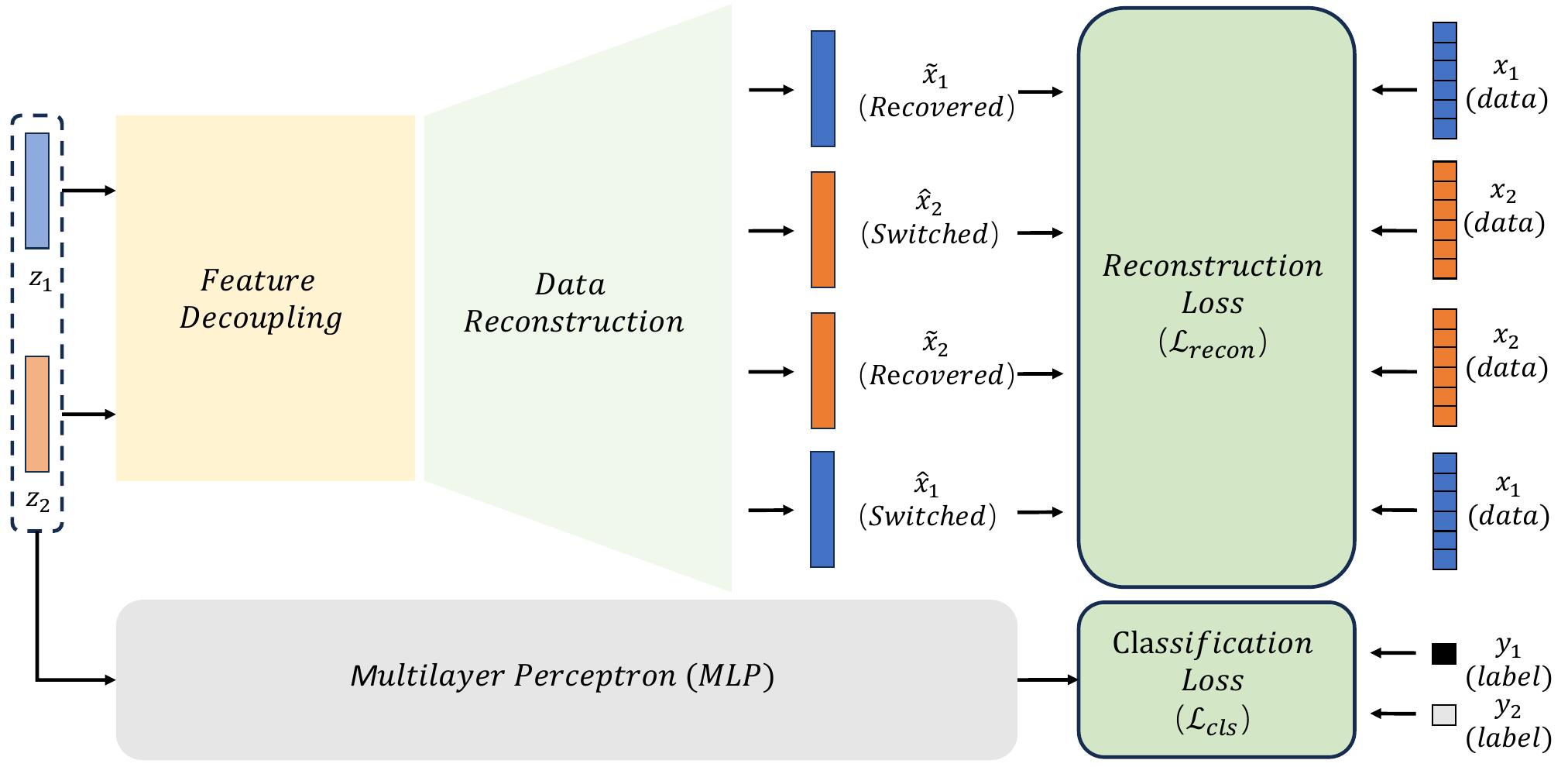}
\end{center}
\vspace*{-3mm} 
  \caption{Block diagram of the proposed pre-training framework with labels. (1) Supervised learning: latent feature vectors $z_1$ and $z_2$ are passed through a multi-layer perceptron (MLP) to predict labels. The cross-entropy loss is computed based on the predicted labels and the true labels. (2) Self-supervised learning: reconstructed (recovered and switched) data and original encoded data are used for computing the mean square error (MSE).}
  \vspace*{-3mm} 
\label{fig: semi}
\end{figure*}

\subsection{Self-supervised Learning}
\label{sec: ssl}
Self-supervised learning of SwitchTab aims to learn informative representations from unlabeled data (Algorithm~\ref{algo:DA}), which is described in Figure~\ref{fig: ssl}. For each of the two data samples, $x_1$ and $x_2$, we apply feature corruption to obtain corrupted data. We encode them using an encoder, $f$, resulting in two feature vectors, $z_1$ and $z_2$. Importantly, we decouple these two feature vectors using two types of projectors, $p_m$ and $p_s$, which extract switchable mutual information among the data samples and salient information that is unique to each individual data sample, respectively. Through this decoupling process, we obtain the salient feature vectors, $s_1$ and $s_2$, and the mutual feature vectors, $m_1$ and $m_2$, for $x_1$ and $x_2$, respectively.

Notably, the mutual features should be shared and switchable between two samples. In other words, the concatenated feature vector of $s_1 \oplus m_1$ should exhibit no discernible difference compared to $s_1 \oplus m_2$. Consequently, it is expected that not only should the decoded data $\tilde{x}_1$ (recovered) from $s_1 \oplus m_1$ be highly similar to $x_1$, but also the decoded data $\hat{x}_1$ (switched) from the concatenated feature vector of $s_1\oplus m_2$ should demonstrate a comparable level of similarity. Likewise, we anticipate both $\tilde{x}_2$ (recovered) and $\hat{x}_2$ (switched) to resemble $x_2$ as much. Therefore, we define the loss function $\mathcal{L}_{\it{self}}=\mathcal{L}_{\it{recon}}$ as reconstruction loss by:
\begin{align}
     \mathcal{L}_{\it{recon}} = \underbrace{\frac{1}{M}\sum_{j=1}^{M} (x_{1_j} - \hat{x}_{1_j})^2 + \frac{1}{M}\sum_{j=1}^{M} (x_{2_j} - \hat{x}_{2_j})^2}_{\text{switched}} \nonumber \\ 
     + \underbrace{\frac{1}{M}\sum_{j=1}^{M} (x_{1_j} - \tilde{x}_{1_j})^2 + \frac{1}{M}\sum_{j=1}^{M} (x_{2_j} - \tilde{x}_{2_j})^2}_{\text{recovered}} .
\end{align}


\subsection{Pre-training with Labels}
\label{sec: semi}
We further improve the pre-training process by taking advantage of labeled data, as shown in Figure~\ref{fig: semi}. With labels introduced, we pose additional constraints to the encoded embeddings $z_1$ and $z_2$ for label prediction and compute the prediction loss (illustrated by classification loss $\mathcal{L}_{\it{cls}}$ through the context). To be specific, $z_1$ and $z_2$ are fed to the same multi-layer perceptron (MLP) that maps from the embedding space to the label space. During the optimization stage, we combine the prediction loss with $\mathcal{L}_{\it{recon}}$ above to update the parameters in the framework. Formally, we define the loss function $\mathcal{L}_{\it{total}}$ for two samples $x_{1}$ and $x_{2}$ as follow:
\begin{align}
        \mathcal{L}_{\it{total}} =  \mathcal{L}_{\it{recon}} 
        + \alpha * \mathcal{L}_{\it{cls}},
\end{align}
where $\alpha$ is used to balance the classification loss and reconstruction loss and set to 1 as default. To illustrate, the cross-entropy loss for classification task is defined as:
\begin{align}
    \mathcal{L}_{\it{cls}} = -\left( y_{1} \log(\hat{y}_{1}) + y_{2} \log(\hat{y}_{2}) \right),
\end{align}
where $\hat{y}_{1}$ and $\hat{y}_{2}$ are predicted labels, i.e., $\hat{y}_{1}= \text{MLP}(z_1)$ and $\hat{y}_{2} = \text{MLP}(z_2)$. For regression tasks, rooted mean squared error (RMSE) will replace the cross-entropy loss.

\begin{table*}\renewcommand{\arraystretch}{1.5}\addtolength{\tabcolsep}{0pt}
    \centering
    \resizebox{1\textwidth}{!}{
    \begin{tabular}{lccccccccccc}
    \toprule
    \textbf{Dataset size}   & 48842 & 65196 & 83733 & 98050 & 108000 & 500000 & 518012 & 20640 & 515345 & 709877 & 1200192 \\   
    \textbf{Feature size}   & 14 & 27 & 54 & 28 & 128 & 2000 & 54 & 8 & 90 & 699 & 136 \\   
    \midrule
    \textbf{Method/Dataset} & \textbf{AD $\uparrow$} & \textbf{HE $\uparrow$} & \textbf{JA $\uparrow$} & \textbf{HI $\uparrow$} & \textbf{AL $\uparrow$} & \textbf{EP $\uparrow$} & \textbf{CO $\uparrow$} & \textbf{CA $\downarrow$} & \textbf{YE $\downarrow$} & \textbf{YA $\downarrow$} & \textbf{MI $\downarrow$} \\
    \midrule
    TabNet   & 0.850 & 0.378 & 0.723 & 0.719 & 0.954 & 0.8896 & 0.957 & 0.510 & 8.909 & 0.823 & 0.751 \\
    SNN      & 0.854 & 0.373 & 0.719 & 0.722 & 0.954 & 0.8975 & 0.961 & 0.493 & 8.895 & 0.761 & 0.751 \\
    AutoInt  & 0.859 & 0.372 & 0.721 & 0.725 & 0.945 & 0.8949 & 0.934 & 0.474 & 8.882 & 0.768 & 0.750 \\
    MLP      & 0.852 & 0.383 & 0.723 & 0.723 & 0.954 & 0.8977 & 0.962 & 0.499 & 8.853 & 0.757 & 0.747 \\
    DCN2     & 0.853 & 0.385 & 0.723 & 0.723 & 0.955 & 0.8977 & 0.965 & 0.484 & 8.890 & 0.757 & 0.749 \\
    NODE     & 0.858 & 0.359 & 0.726 & 0.726 & 0.918 & 0.8958 & 0.985 & 0.464 & \textbf{8.784} & 0.753 & 0.745 \\
    ResNet   & 0.854 & \textbf{0.396} & 0.727 & 0.727 & \textbf{0.963} & 0.8969 & 0.964 & 0.486 & 8.846 & 0.757 & 0.748 \\
    FT-Transormer     & 0.859 & 0.391 & {0.729} & 0.729 & 0.960 & {0.8982} & 0.970 & 0.459 &  8.855 & 0.756 & 0.746  \\
    \midrule
    XGBoost  & 0.874 & 0.377 & 0.724 & 0.728 & 0.924 & 0.8799 & 0.964 & 0.431 & {8.819} & \textbf{0.732} & \textbf{0.742} \\
    CatBoost & 0.873 & 0.388 & 0.727 & {0.729} & 0.948 & 0.8893 & 0.950 & \textbf{0.423} & 8.837 & 0.740 & 0.743 \\
    
    \midrule
    SwitchTab (Self-Sup.) & {0.867} & 0.387 & {0.726} & {0.724} & 0.942 & {0.8928} & {0.971} & 0.452 & 8.857 & 0.755 & {0.751} \\
    SwitchTab & \textbf{0.881} & 0.389 & \textbf{0.731} & \textbf{0.733} & 0.951 & \textbf{0.8987} & \textbf{0.989} & 0.442 & 8.822 & 0.744 & \textbf{0.742} \\
    \bottomrule
    \end{tabular}
    }
    \caption{Comparison of different methods on the previous benchmark. For each dataset, the best results are shown in \textbf{Bold}. Reported results are averaged over three trials. Notations: $\downarrow\sim$ RMSE for regression task, $\uparrow\sim$ accuracy for classification task.}
    \label{tab: benchmark}
      \vspace*{-5mm} 
\end{table*}

\subsection{Downstream Fine-tuning}
In line with the established paradigm of representation learning \cite{he2020momentum, chen2020improved, chen2020simple, bahri2021scarf}, we perform the end-to-end fine-tuning of the pre-trained encoder from SwitchTab using the complete set of labeled data. Specifically, we incorporate the encoder $f$ with an additional linear layer, unlocking all its parameters and adapting them for the downstream supervised tasks. 

Another avenue to leverage the advantages of our framework lies in harnessing the salient feature vector $s$ as a plug-and-play embedding. By concatenating $s$ with its original feature vector $x$, we construct enriched data sample vector denoted as $x_{concat}=x \oplus s$. This method effectively highlights the distinct characteristics within the data which facilitates the establishment of a clear decision boundary. As a result, we anticipate noticeable enhancements in classification tasks when utilizing $x_{concat}$ as the input for a traditional model like XGBoost.

\section{Experiments and Results}

In this section, we present the results of our comprehensive experiments conducted on various datasets to demonstrate the effectiveness of SwitchTab. The section is divided into two parts. In the first part, we provide preliminary information about the experiments, including the datasets, data preprocessing, model architectures, and training details, aiming to ensure transparency and reproducibility.

In the second part, we evaluate the performance of our proposed method from two distinct perspectives. First, we compare SwitchTab against mainstream deep learning and traditional models using standard benchmarks from \cite{gorishniy2021revisiting} and additional datasets to establish a more comprehensive performance assessment of SwitchTab. Secondly, we showcase the versatility of SwitchTab by demonstrating the utilization of salient features as plug-and-play embeddings across various traditional models, including XGBoost, Random Forest, and LightGBM. The plug-and-play strategy allows us to enhance the traditional models performance effortlessly and without additional complexity.

\begin{table*}\renewcommand{\arraystretch}{1.5}\addtolength{\tabcolsep}{0pt}
    \centering
    \resizebox{1\textwidth}{!}{
    \begin{tabular}{l c>{\columncolor[gray]{0.95}}c>{\columncolor[gray]{0.8}}c | c>{\columncolor[gray]{0.95}}c>{\columncolor[gray]{0.8}}c |c>{\columncolor[gray]{0.95}}c>{\columncolor[gray]{0.8}}c |c>{\columncolor[gray]{0.95}}c>{\columncolor[gray]{0.8}}c |c>{\columncolor[gray]{0.95}}c>{\columncolor[gray]{0.8}}c |c>{\columncolor[gray]{0.95}}c>{\columncolor[gray]{0.8}}c |c>{\columncolor[gray]{0.95}}c>{\columncolor[gray]{0.8}}c  }
    \toprule
    \textbf{Dataset size}   & \multicolumn{3}{c}{45211} & \multicolumn{3}{c}{7043} & \multicolumn{3}{c}{452} & \multicolumn{3}{c}{200} & \multicolumn{3}{c}{12330} & \multicolumn{3}{c}{58310} & \multicolumn{3}{c}{518012} \\   
    \textbf{Feature size}   & \multicolumn{3}{c}{16} & \multicolumn{3}{c}{20} & \multicolumn{3}{c}{226} & \multicolumn{3}{c}{783} & \multicolumn{3}{c}{17} & \multicolumn{3}{c}{147} & \multicolumn{3}{c}{54}  \\   
    \midrule
    \textbf{Dataset} & \multicolumn{3}{c}{\textbf{BK}} & \multicolumn{3}{c}{\textbf{BC}} & \multicolumn{3}{c}{\textbf{AT}} & \multicolumn{3}{c}{\textbf{AR}} & \multicolumn{3}{c}{\textbf{SH}} & \multicolumn{3}{c}{\textbf{VO}$\bigstar$}  & \multicolumn{3}{c}{\textbf{MN}$\bigstar$} \\
    \midrule
    \textbf{Raw Feature ($x$)}  &\cmark &   &\cmark &\cmark &   &\cmark &\cmark &   &\cmark &\cmark &   &\cmark &\cmark &   &\cmark &\cmark &   &\cmark &\cmark &   &\cmark   \\
    \textbf{Salient Feature ($s$)}  &  &\cmark &\cmark &  &\cmark &\cmark &  &\cmark &\cmark &  &\cmark &\cmark &  &\cmark &\cmark &  &\cmark &\cmark &  &\cmark &\cmark \\
    \midrule
    Logistic Reg.   & 0.907 &0.910 & 0.918      &0.892  &0.894 &0.902       & 0.862&0.862  &0.869     & 0.916 &0.915 &0.922         & 0.870 &0.871 &  0.882      & 0.539&0.545 &0.551        & 0.899 &0.907 &0.921     \\
    
    Random Forest   & 0.891 &0.895 & 0.902      &0.879  &0.880 &0.899       & 0.850&0.853  &0.885     & 0.809 &0.810 &0.846         & 0.929 &0.931& 0.933        &0.663  &0.669 &0.672       &0.938 &0.940 & 0.945 \\
    
    XGboost         & 0.929 &0.929 & 0.938      &0.906  &0.907 &0.912       & 0.870&0.872  &0.904     & 0.824 &0.828 &0.843      &0.925 &0.924&0.931             &0.690  &0.691 &0.693       &0.958 &0.961  &0.964 \\
    
    LightGBM        & 0.939 &0.939 & 0.942      &0.910  &0.910 &0.915       & 0.887&0.889  &0.903     & 0.821 &0.826 &0.831      &0.932 &0.933&0.944             & 0.679 &0.682 &0.686       &0.952 &0.955 &0.963 \\
    
    CatBoost        & 0.925 &0.928 & 0.937      &0.912  &0.910 &0.919       & 0.879&0.880  &0.899     & 0.825 &0.828 &0.877      &0.931 &0.934&0.942             &0.664  &0.671 &0.682       &0.956 &0.958 &0.968  \\
    
    MLP             & 0.915 &0.917 & 0.923      &0.892  &0.895 &0.902       & 0.902&0.905  &0.912     & 0.903 &0.904 &0.908          &0.887 &0.891&0.910         &0.631  &0.633 &0.642       &0.939 &0.941 &0.948 \\
     \midrule   
    VIME            & 0.766 & - & -    &0.510& - & -   & 0.653 & - & -        & 0.610 & - & -      &0.744 & - & -      &0.623 & - & -  &0.958  & - & -\\
    TabNet          & 0.918 & - & -    &0.796 & - & -   & 0.521 & - & -           &0.541  & - & -      &0.914 & - & -      &0.568 & - & -  &0.968  & - & - \\
    TabTransformer       & 0.913 & - & -    &0.817 & - & -   & 0.700 & - & -           &0.868  & - & -      &0.927 & - & -      &0.580 & - & - &0.887  & - & -\\
    SAINT           & 0.933 & - & -    &0.847 & - & -   &  \textbf{0.941}& - & -   & 0.910 & - & -    &0.931 & - & -      &0.701 & - & -  &0.977  & - & -  \\

    {ReConTab} &{0.929}   & - & -  &{0.913}   & - & -     &{0.907} & - & - & {0.918}  & - & - &{0.931}   & - & - &{0.680}   & - & - &{0.968}   & - & - \\

    \midrule
    SwitchTab(Self-Sup.) &{0.917}   & - & -  &{0.903}   & - & -      &{0.900} & - & - & {0.904}  & - & - &{0.931}   & - & - &{0.629}   & - & - &{0.969}   & - & - \\
    
    \textbf{SwitchTab} &\textbf{0.942}   & - & -  &\textbf{0.923}   & - & -     &{0.928} & - & - & \textbf{0.922}  & - & - &\textbf{0.958}   & - & - &\textbf{0.708}   & - & - &\textbf{0.982}   & - & - \\
    \bottomrule
    \multicolumn{22}{l}{\normalsize $``-"$ indicates the experiments are not applicable for the corresponding methods to demonstrate the benefits of plug-and-play embeddings.} \\
    \end{tabular}
    }
    \caption{Comparison of different methods on classification task. For each method, we report three categories 1) raw features only, 2) salient features only, 3) plug and play using salient features. The best results are shown in \textbf{Bold}. Columns added with $\bigstar$ are multi-class classification tasks, reporting accuracy. The other results of binary classification tasks are evaluated with AUC.}
  \vspace*{-5mm} 
    \label{tab: further}
\end{table*}

\subsection{Preliminaries for Experiments}

\subsubsection{Datasets.} We first evaluate the performance of SwitchTab on a standard benchmark from \cite{gorishniy2021revisiting}. Concretely, the datasets include: California Housing (CA) \cite{pace1997sparse}, Adult (AD) \cite{kohavi1996scaling}, Helena (HE) \cite{guyon2019analysis}, Jannis (JA) \cite{guyon2019analysis}, Higgs (HI) \cite{baldi2014searching}, ALOI (AL) \cite{geusebroek2005amsterdam}, Epsilon (EP) \cite{yuan2011improved}, Year (YE) \cite{bertin2011million}, Covertype (CO) \cite{blackard1999comparative}, Yahoo (YA) \cite{chapelle2011yahoo}, Microsoft (MI) \cite{qin2013introducing}.

Besides the standard benchmarks, there is also another set of popular datasets used by recent work \cite{somepalli2021saint}, including Bank (BK) \cite{moro2014data}, Blastchar (BC) \cite{ouktelco}, Arrhythmia (AT) \cite{liu2008isolation,ouktelco}, Arcene (AR) \cite{asuncion2007uci}, Shoppers (SH) \cite{sakar2019real}, Volkert (VO) \cite{automlchallenges} and MNIST (MN) \cite{xiao2017fashion}. 

\subsubsection{Preprocessing of Datasets.}
We represent categorical features using a backward difference encoder \cite{potdar2017comparative}. Regarding missing data, we discard any features that are missing for all samples. For the remaining missing values, we employ imputation strategies based on the feature type. Numerical features are imputed using the mean value, while categorical features are filled with the most frequent category found within the dataset. Furthermore, we ensure uniformity by scaling the dataset using a Min-Max scaler. When dealing with image-based data, we flatten them into vectors, thus treating them as tabular data, following the approach established in prior works \cite{yoon2020vime, somepalli2021saint}. 

\subsubsection{Model Architectures.} 
For feature corruption, we uniformly sample a subset of features for each sample to generate a corrupted view at a fixed corruption ratio of 0.3. For the encoder $f$, we employ a three-layer transformer with two heads. The input and output sizes of the encoder are always aligned with the feature size of the input. Both projectors $p_s$ and $p_m$ consist of one linear layer, followed by a sigmoid activation function. Additionally, the decoder $d$ remains a one-layer network with a sigmoid activation function. During the pre-training stage with labels, we introduce an additional one-layer network for prediction. In the downstream fine-tuning stage, we append a linear layer after the encoder $f$ to accommodate classification or regression tasks.

\subsubsection{Training Details.} Importantly, we maintain consistent settings throughout the evaluation of SwitchTab. Although further gains might be attainable with further exploration of hyperparameters, we intentionally refrain from doing so to ensure the proposed approach can be easily generalized across diverse datasets and domains. For all the pre-training, we train all models for 1000 epochs with the default batch size of 128. We use the RMSprop optimizer \cite{hinton2012neural} with an initial learning rate set as 0.0003. During the fine-tuning stage, we set the maximum epochs as 200. Adam optimizer with a learning rate of 0.001 is used.

\begin{table}
    \centering
    \resizebox{0.47\textwidth}{!}{
    \begin{tabular}{l c | c | c | c | c | c | c }
    \toprule
    \textbf{\makecell[c]{Dataset }} & \multicolumn{1}{c}{\textbf{BK}} & \multicolumn{1}{c}{\textbf{BC}} & \multicolumn{1}{c}{\textbf{AT}} & \multicolumn{1}{c}{\textbf{AR}} & \multicolumn{1}{c}{\textbf{SH}} & \multicolumn{1}{c}{\textbf{VO}$\bigstar$}  & \multicolumn{1}{c}{\textbf{MN}$\bigstar$} \\
    \midrule
    \textbf{\makecell[c]{SwitchTab\\ (No Switching) }} & 0.918 & 0.909 & 0.902  & 0.896 & 0.912 & 0.689 & 0.968\\
    \midrule
    \textbf{\makecell[c]{SwitchTab }}   &\textbf{0.942}   &\textbf{0.923}    &  \textbf{0.928}   & \textbf{0.922}    &\textbf{0.958}  &\textbf{0.708}      &\textbf{0.982}   \\
    \bottomrule
    \end{tabular}
    }
    \caption{Ablation of model performance w.r.t the switching process. Columns added with $\bigstar$ are multi-class classification tasks, reporting their accuracy. The other results of binary classification tasks are evaluated with AUC.}
    \label{tab: ablation1}
    \vspace*{-5mm} 
\end{table}

\subsection{Results on Previous Benchmarks}

We conduct a comprehensive performance comparison of SwitchTab with different methods across 11 datasets from previous benchmarks, as shown in Table~\ref{tab: benchmark}. To ensure a fair and direct comparison, we report the accuracy of the classification tasks, following the metrics employed in previous studies. It is worth noting that we meticulously fine-tuned the results in accordance with the established paradigm \cite{kolesnikov2019revisiting}. Upon analyzing the results, we find that SwitchTab consistently achieves optimal or near-optimal performance in most of the classification tasks. These outcomes underscore the effectiveness and superiority of SwitchTab in representation learning for classification scenarios. However, in regression tasks, we observe that traditional methods like XGBoost or CatBoost still dominate and achieve the best results. Nonetheless, SwitchTab remains highly competitive and outperforms various deep learning approaches in these regression scenarios. We report the averaged results over 10 random seeds. 

\subsection{Results on Additional Public Datasets}

Beyond the previous benchmarks, we continue the performance comparisons on additional public datasets and summarize the results in Table~\ref{tab: further}. The results encompass evaluations using both traditional models and more recent deep learning techniques. In the majority of cases, SwitchTab showcases remarkable improvements, surpassing all baseline methods and reinforcing its superiority across diverse datasets and scenarios. However, it is essential to acknowledge that on the dataset AT, SwitchTab achieved sub-optimal results when compared to the baselines. This observation aligns with previous research conclusions that the tabular domain poses unique challenges where no single method universally dominates \cite{gorishniy2021revisiting}. Nevertheless, this outcome merits further investigation to discern the specific factors contributing to this variation in performance.

\subsection{Plug-and-Play Embeddings}
As mentioned earlier, SwitchTab excels in effectively extracting salient features which could significantly influence the decision boundaries for classification tasks. In the plug-and-play setting, our experiment results demonstrate that these salient features have immense value when integrated with original data as additional features. Notably, the performance of all traditional methods can be boosted, improving the evaluation metrics (e.g., AUC) from $0.5\%$ to $3.5\%$ (in absolute difference) across various datasets, as illustrated in the dark gray columns Table~\ref{tab: further}. Meanwhile, we also report results when using only the salient features as input. While the improvement is relatively marginal, it aligns with our expectations. The absence of mutual information in this scenario leads to a less substantial performance boost.

\begin{figure}[t!]
\begin{center}
    \includegraphics[width=0.85\linewidth]{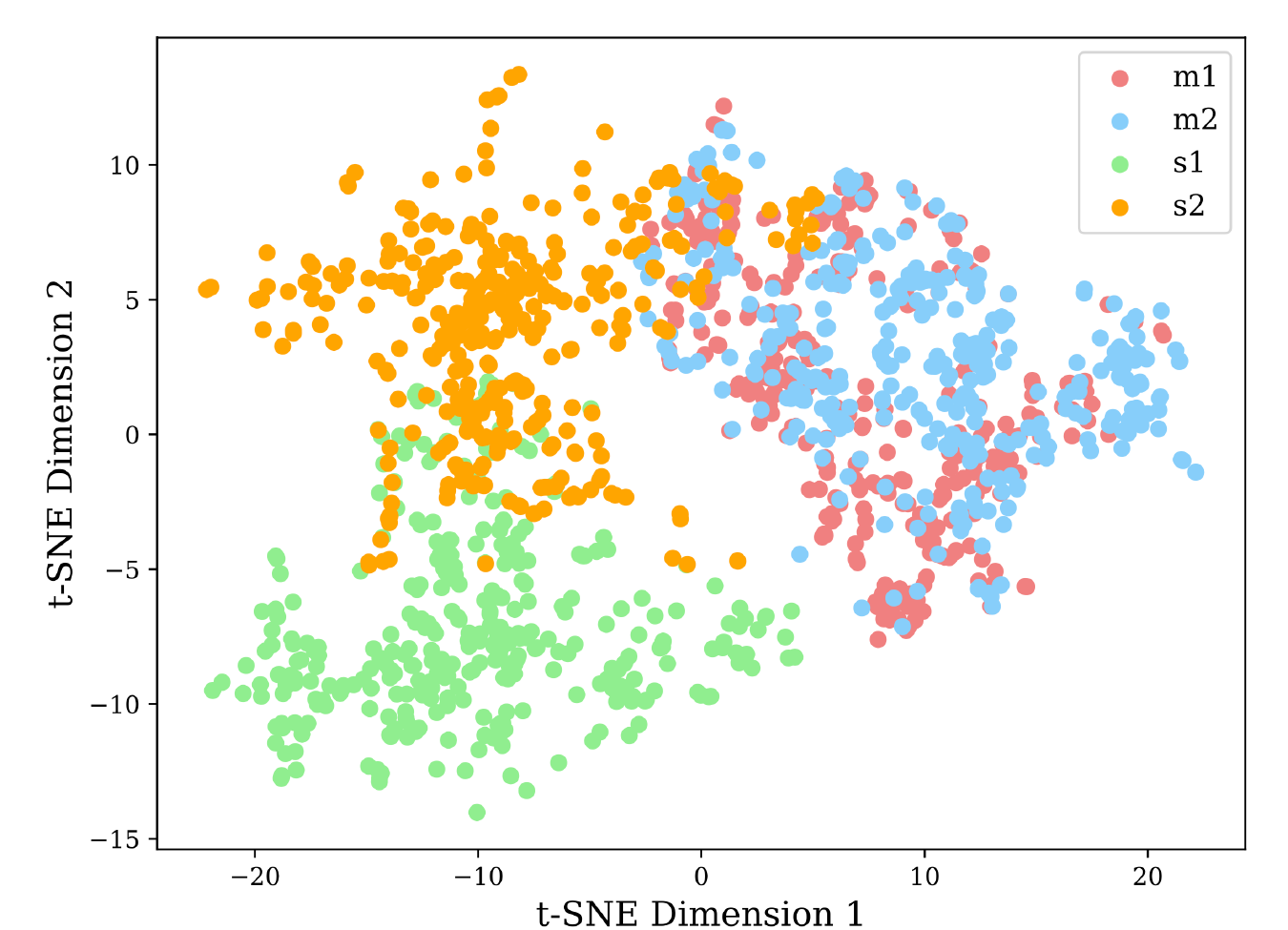}
\end{center}
  \caption{t-SNE visualization of mutual and salient features in two-dimensional space.}
  \vspace*{-5mm} 
\label{fig: visual}
\end{figure}

\subsection{Visualization and Discussions}
In this section, we visualize the features learned by SwitchTab using the BK dataset, which is designed for binary classification tasks. After pre-training, we feed the first batch with data from one class and the second batch with data from the other class, and then visualize the corresponding feature vectors. As shown in Figure~\ref{fig: visual}, the embeddings $m_1$ and $m_2$ from SwitchTab, although extracted from two different classes, heavily overlap with each other. This substantiates the fact that the mutual information is switchable. However, the salient feature $s_1$ and the salient feature $s_2$ are distinctly separated, playing a dominant role in capturing the unique properties of each class and decisively contributing to the classification boundaries.

\subsection{Ablation Studies}
In this section, we investigate essential modules of SwitchTab, including the importance of the switching process, the feature corruption rate and the computation cost. We use all of the datasets in Table~\ref{tab: further}, with all the same data preprocessing and optimization strategies.

\subsubsection{Contribution of Switching Process.}
To demonstrate that the superior performance of the proposed model directly results from the critical switching process, we report the results with and without reconstructing the concatenated features from switched pairs, i.e., $(s_1, m_2)$ and $(s_2, m_1)$, keeping the feature corruption ratio at 0.3 for all experiments. Notably, without the switching mechanism, the framework deteriorates to a simpler auto-encoder structure and results in obvious drop in evaluation metrics (e.g., AUC) in Table~\ref{tab: ablation1}. 

\subsubsection{Feature Corruption Ratio.} 
We also explore the optimal feature corruption ratio in Table~\ref{tab: ablation2}. Through extensive analysis, we find that the optimal corruption ratio is approximately 0.3. Therefore, we adopt this value as the default for all previously reported experiments. However, it is essential to emphasize that this selected ratio may not be consistently optimal for each dataset. We also observe that datasets with higher feature dimensions, such as AR or VO, tend to benefit from larger corruption ratios, since they are more likely to have redundant features. This observation is aligned with previous conclusions on tabular data from \cite{grinsztajn2022tree}. Conversely, for datasets with low-dimensional features such as BC, smaller corruption ratios could also yield superior results in our experiments.

    
    

\begin{table}\renewcommand{\arraystretch}{0.9}\addtolength{\tabcolsep}{0pt}
    \centering
    \small
    \resizebox{0.47\textwidth}{!}{
    \begin{tabular}{l c | c | c | c | c | c | c }
    \toprule
    \textbf{Ratio} & \multicolumn{1}{c}{\textbf{0.0}} & \multicolumn{1}{c}{\textbf{0.1}} & \multicolumn{1}{c}{\textbf{0.2}} & \multicolumn{1}{c}{\textbf{0.3}} & \multicolumn{1}{c}{\textbf{0.4}} & \multicolumn{1}{c}{\textbf{0.5}}  & \multicolumn{1}{c}{\textbf{0.6}} \\
    \midrule
    \textbf{BK}  & 0.927 &{0.938}  & 0.940 &\textbf{0.942}   & 0.932 &  {0.903}  & 0.898    \\
    \midrule
    \textbf{BC} & 0.911  &{0.920}  & \textbf{0.923} &\textbf{0.923}   & 0.917 &  {0.910}  & 0.902   \\
    \midrule
    \textbf{AT} & 0.916  &{0.922}  & 0.925 &\textbf{0.928}   & 0.927 &  {0.920}  & 0.913   \\
    \midrule
    \textbf{AR} & 0.913  & {0.915}  & 0.918 & {0.922}   & \textbf{0.925} &  {0.920}  & 0.914    \\
    \midrule
    \textbf{SH} & 0.948  &{0.956}  & 0.956 &\textbf{0.958}   & 0.947 &  {0.934}  & 0.922     \\
    \textbf{\textbf{VO}$\bigstar$} & 0.683  &{0.694}  & 0.699 &{0.708}   & \textbf{0.709} &  {0.700}  & 0.692    \\
    \textbf{\textbf{MN}$\bigstar$} & 0.969  &{0.971}  & 0.977 &\textbf{0.982}   & 0.978 &  {0.966}  & 0.957    \\
    \bottomrule
    \end{tabular}
    }
    \caption{Ablation of feature corruption ratio. Multi-class classification tasks with $\bigstar$ are reporting accuracy. The other binary classification tasks are evaluated with AUC.}
    \label{tab: ablation2}
      \vspace*{-5mm} 
\end{table}

\section{Conclusion}
Motivated by the profound success of representation learning in computation vision and natural language processing domains, we want to extend this success to tabular data domain. Differentiating from other related studies to address this issue from a contrastive learning perspective, we introduce SwitchTab, a novel pre-training framework for representation learning from the perspective of generative models. The learned embeddings from SwitchTab could not only achieve superior performance on downstream tasks but also represent a distinguishable salient feature space that can enhance a broader range of traditional methods as plug-and-play embeddings. We firmly believe that this work constitutes a critical step towards achieving more representative, explainable, and structured representations for tabular data.




\bibliography{aaai24}

\end{document}